\documentclass[11pt]{article} 
\usepackage[]{ACL2023}
\usepackage{times}  
\usepackage{helvet}  
\usepackage{courier}  
\usepackage{url}  
\usepackage{graphicx} 
\usepackage{natbib}  
\usepackage{caption} 
\usepackage{algorithm}
\usepackage{algorithmic}

\usepackage{tabularx}
\usepackage{bm}
\usepackage{amsmath}
\usepackage{mathrsfs}
\usepackage{multirow}
\usepackage{booktabs}
\usepackage{tikz}
\usepackage{tikz-qtree}
\usepackage{color}
\usepackage{xcolor}
\usepackage{makecell}
\usepackage[switch]{lineno}
\usepackage{enumitem}
\usepackage{makecell}
\usepackage{bbding}
\usepackage{subfigure}

\usepackage{newfloat}
\usepackage{listings}
\DeclareCaptionStyle{ruled}{labelfont=normalfont,labelsep=colon,strut=off} 

\floatstyle{ruled}
\newfloat{listing}{tb}{lst}{}
\floatname{listing}{Listing}
\title{A Simple yet Effective Self-Debiasing Framework for Transformer Models}

\author{Xiaoyue Wang$^{a,*}$, Lijie Wang$^{b,*}$, Xin Liu$^c$, Suhang Wu$^a$,  Jinsong Su$^{a,\dagger}$, Hua Wu$^b$ \\
$^a$School of Informatics, Xiamen University, Xiamen 361005, China,\\ 
$^b$Baidu Inc., Beijing 100085, China,\\
$^c$University of Michigan, Ann Arbor 48104, USA \\
\texttt{xiaoyuewang@stu.xmu.edu.cn, wanglijie@baidu.com, jssu@xmu.edu.cn}}

\begin{document}

\maketitle

\begin{abstract}

Current Transformer-based natural language understanding (NLU) models heavily rely on dataset biases, while failing to handle real-world out-of-distribution (OOD) instances.
Many methods have been proposed to deal with this issue, but they ignore the fact that the features learned in different layers of Transformer-based NLU models are different. In this paper, we first conduct preliminary studies to obtain two conclusions: 1) both low- and high-layer sentence representations encode common biased features during training; 2) the low-layer sentence representations encode fewer unbiased features than the high-layer ones. Based on these conclusions, we propose a simple yet effective self-debiasing framework for Transformer-based NLU models. Concretely, we first stack a classifier on a selected low layer. Then, we introduce a residual connection that feeds the low-layer sentence representation to the top-layer classifier. In this way, the top-layer sentence representation will be trained to ignore the common biased features encoded by the low-layer sentence representation and focus on task-relevant unbiased features. During inference, we remove the residual connection and directly use the top-layer sentence representation to make predictions.
Extensive experiments and in-depth analyses on NLU tasks show that our framework performs better than several competitive baselines, achieving a new SOTA on all OOD test sets.\footnote{We release our code at \url{https://github.com/bigcat2333/DeRC}.}

\end{abstract}
\renewcommand{\thefootnote}{\fnsymbol{footnote}}
\footnotetext[1]{Both authors contributed equally to this work.}
\footnotetext[2]{Corresponding author.}
\section{Introduction}

Recently, Transformer-based models have achieved competitive performance on various NLU benchmarks \cite{ wang-etal-2018-glue, devlin-etal-2019-bert}. However, many studies show that these models tend to directly exploit biased features as shortcuts to make predictions without understanding the semantics of input texts \cite{gururangan-etal-2018-annotation, hans2019, geirhos2020shortcut}. As a result, this behavior leads to the low generalizability and poor robustness of these models on OOD instances \cite{zellers-etal-2018-swag}. For example, on the PAWS \cite{zhang-etal-2019-paws}, which is the OOD test set for Quora Question
Pairs (QQP) dataset\footnote{https://www.kaggle.com/c/quora-question-pairs}. The commonly-used BERT-based model \cite{devlin-etal-2019-bert} does not achieve the expected results, as shown in Table \ref{tab:case_intro}. 

\begin{table}[tb]
\small
\center
\renewcommand{\arraystretch}{1.1}
\begin{tabularx}{0.45\textwidth}{ l }
\toprule[1pt]
\makecell[l]{\textbf{Sentence 1}:  \textit{`` Captain '' was broken up in 1762.} \\ \textbf{Sentence 2}:  \textit{ `` Captain '' was rolled up in 1762.} \\  \textbf{Golden Label}:  \textit{non-duplicate} \\
\textbf{Predicted Label}:  \textit{\textcolor{red} {duplicate}}} \\
\hline
\makecell[l]{\textbf{Sentence 1}:  \textit{Is there a tutorial on how to use Quora?} \\ \textbf{Sentence 2}:  \textit{How do I start using Quora?} \\ \textbf{Golden Label}:  \textit{duplicate} \\
\textbf{Predicted Label}: \textit{\textcolor{red}{non-duplicate}}} \\
\bottomrule[1pt]
\end{tabularx}
\caption{Two instances from PAWS \cite{zhang-etal-2019-paws}. Both instances contain biased features, which make the dominant model \cite{devlin-etal-2019-bert} unable to predict the relationship between their sentences correctly. In the first instance, its two sentences contain a high proportion of overlapping words, which convey different meanings. The second instance is a paraphrase sentence pair, while its two sentences contain a limited number of overlapping words. }
\label{tab:case_intro}
\end{table}

To deal with this issue, many model-agnostic debiasing methods have been proposed, which mainly involve two steps. The first step is identifying biased training instances, of which predictions are heavily influenced by biased features, via data analysis, researchers' task-specific insights \cite{gururangan-etal-2018-annotation, poliak-etal-2018-hypothesis, tsuchiya-2018-performance, feversymm2019} 
or bias-only models \cite{utama-etal-2020-mind,utama-etal-2020-towards, ghaddar-etal-2021-end}. 
The second step is employing various methods to down-weight the importance of biased training instances, such as example re-weighting \cite{feversymm2019, karimi-mahabadi-etal-2020-end}, confidence regularization \cite{utama-etal-2020-mind} and model ensemble \cite{clark-etal-2019-dont, he-etal-2019-unlearn, karimi-mahabadi-etal-2020-end}.

Despite their success, most studies consider NLU models as black-box systems, ignoring that different layers of Transformer-based NLU model learn different features. As analyzed in previous studies \cite{bert-structure2019, tenney-etal-2019-bert}, Transformer-based pre-trained language models are able to effectively capture rich linguistic knowledge, with surface features in low layers, syntactic features in middle layers, and semantic features in high layers. Thus, two questions naturally arise: 1) Are there differences in features learned in terms of bias by different layers, i.e., biased and unbiased feature learning? 2) If so, can we leverage these differences to alleviate biased feature learning?

To answer the first question, we conduct preliminary studies to explore feature learning in different layers of Transformer-based NLU models. Specifically, following \citet{du-etal-2021-towards}, we first identify biased and anti-biased training instances from the training set, and extract biased and anti-biased validation instances from the validation set. Then, we stack a classifier on the sentence representation of each Transformer layer. 
Afterwards, we analyze the feature learning of different layers from both model training and prediction perspectives.
Experimental analyses show that 1) the low- and high-layer sentence representations encode common biased features, and 2) the low-layer sentence representations encode fewer unbiased features than the high-layer ones.

Based on the above analyses, we propose a self-debiasing framework for Transformer-based NLU models. 
Concretely, we first add a classifier on a selected low layer to encourage the low-layer sentence representation to encode more common biased features during training, which are also encoded in the high-layer classifier. Then, we introduce a residual connection \cite{he2016deep} that feeds the low-layer sentence representation to the top-layer classifier. 
In this way, 
the top-layer sentence representation is encouraged to ignore the common biased features and pay attention to task-relevant unbiased features.
Note that we remove the residual connection during inference and directly use the top-layer sentence representation to make predictions.

Finally, we conduct experiments on three NLU tasks. Experimental results show that our simple framework not only achieves better performance on the OOD test sets, but also maintains comparable performance on the validation sets, compared with previous methods \cite{utama-etal-2020-mind, utama-etal-2020-towards, du-etal-2021-towards}. Besides, we prove that our framework indeed improves the understanding ability of the model through in-depth analyses. 
\section{Related Work}
Our related works mainly include the studies on identifying biased instances and debiasing methods.

\vspace{-2mm}
\paragraph{Identifying Biased Instances} This task is crucial to the subsequent debiasing methods. In this respect, many researchers first manually characterize the specific types of dataset biases, including word co-occurrence \cite{gururangan-etal-2018-annotation, poliak-etal-2018-hypothesis, tsuchiya-2018-performance, feversymm2019} and lexico-syntactic patterns \cite{snow-etal-2006-effectively, zellers-etal-2019-hellaswag}, and then identify biased instances according to these bias patterns. However, these methods heavily rely on researchers’ intuition and task-specific insights, limiting their applications to various NLU tasks and datasets. 
To deal with this issue, some studies employ various methods to create bias-only models for identifying biased instances, such as using a tiny fraction of training data \cite{utama-etal-2020-towards}, partial inputs \cite{clark-etal-2020-learning, sanh2020learning, karimi-mahabadi-etal-2020-end}, or a simplified model architecture \cite{ghaddar-etal-2021-end}.

\vspace{-2mm}
\paragraph{Debiasing Methods} 
There have been many attempts to reduce dataset biases through various data construction methods, such as adversarial filtering \cite{nie2020adversarial} and human-in-the-loop \cite{lee2021crossaug}. Despite their effectiveness, researchers also show that newly constructed datasets may not cover all biased patterns \cite{sharma-etal-2018-tackling}. Therefore, many researchers resort to various robust algorithms based on their prior knowledge of task-specific biases. In this respect, some studies adopt adversarial learning to remove the hypothesis-only bias from NLI models. For example, \citet{belinkov-etal-2019-adversarial} and \citet{stacey-etal-2020-avoiding} apply the gradient reverse layer \cite{ganin2015unsupervised} to train an external classifier that forces the hypothesis encoder to ignore hypothesis-only biases. A complementary line of studies focus on debiasing models by down-weighting the importance of biased instances during training, such as example re-weighting \cite{feversymm2019, utama-etal-2020-towards}, confidence regularization \cite{utama-etal-2020-mind}, upweighting minority instances
\cite{tu2020empirical,yaghoobzadeh2021increasing}, and model ensemble \cite{clark-etal-2019-dont, karimi-mahabadi-etal-2020-end}. Usually, these methods involve two models, i.e., a bias-only model used to identify biased instances and a robust model learning from anti-biased instances. In addition to the above, very recently \citet{lyu2022feature} use contrastive learning to capture the dynamic influence of biases, then reduce biased features.

Notably, most of previous studies consider models as black-box systems, and use the above two steps to debias models. By contrast, in this work, we explore the debiasing framework based on the internal structure of the model without manual analyses, extra bias-only models or complex hyper-parameter settings.

\section{Feature Learning in Transformer Models}
\label{sec:shortcut-layer}

In this work, we choose BERT \cite{devlin-etal-2019-bert} as our basic model, due to its competitive performance in many NLU tasks \cite{wang-etal-2018-glue}. In this section, we first briefly introduce BERT, and then conduct preliminary studies to analyze the feature learning in different layers of the BERT-based NLU model.

\subsection{Overview of BERT Architecture}
\label{ssec:bert-overview}

BERT stacks $L$ identical layers, each containing a multi-head self-attention sub-layer, an MLP sub-layer, and a residual connection around these two sub-layers, followed by a layer normalization sub-layer.

Note that many studies on representation learning show that BERT can effectively capture rich linguistic knowledge, with different kinds of knowledge in different layers \cite{bert-structure2019, tenney-etal-2019-bert, hewitt-manning-2019-structural, bert-structure2019}. 
The following subsections aim to answer the two research questions shown in the Introduction.

\begin{figure}[h]
\centering
\includegraphics[width=0.43\textwidth]{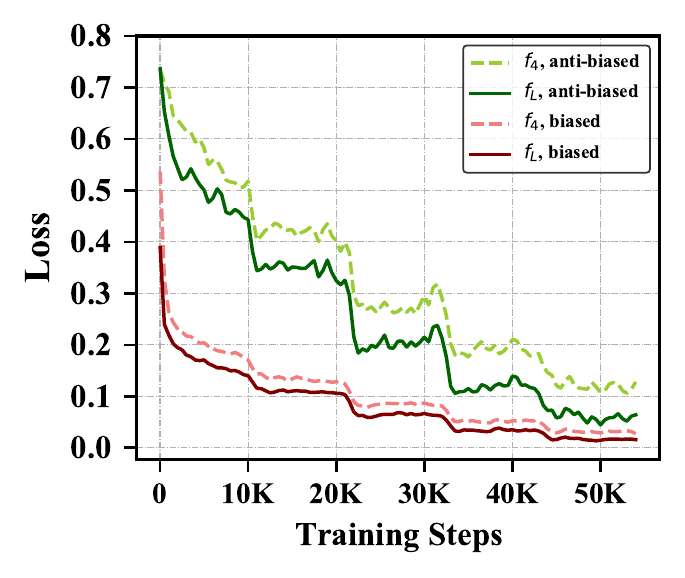}
\vspace{-0.35cm}
\caption{
Training loss curves on biased and anti-biased training instances, where $f_4$ and $f_L$ represent the 4th-layer and top-layer classifier, respectively.
}
\label{fig:training_loss}
\end{figure}

\subsection{Feature Learning in Different Layers of BERT}
\label{ssec:bias-prove-exp}

To answer the above questions, we construct a BERT-based NLU model and equip it with layer-specific classifiers based on sentence representations. Then, we analyze the features learning of these layers from both model training and prediction perspectives.

Previous studies \cite{zhang-etal-2019-paws,geirhos2020shortcut} observed that the lexical overlap of two sentences is a typical biased feature in QQP, and a high lexical overlapping ratio usually co-occurs with some specific labels. Inspired by the above observation, we identify biased and anti-biased training instances from the QQP training set, and biased and anti-biased validation instances from the QQP validation set, respectively, based on the lexical overlapping ratio of each instance.
Concretely, we first calculate the number of overlapping words and divide it by the maximum sentence length. Then, we identify an instance as a biased one if it satisfies the following:
1) its lexical overlapping ratio is greater than 70\% and the label is ``\textit{duplicate}''; 
2) it possesses a ratio less than 30\% and is assigned with an ``\textit{non-duplicate}'' label. 
Conversely, the instance with a ratio greater than 70\% and ``\textit{non-duplicate}'' label, or with a ratio less than 30\% and ``\textit{duplicate}'' label, is considered as an anti-biased instance.

Afterwards, we use the original QQP dataset to train the model and inspect the training losses of different classifiers on the biased and anti-biased training instances, respectively. From Figure \ref{fig:training_loss}\footnote{For the sake of clarity, here we only show the curves of the 4th- and top-layer classifiers, and the curves of other classifiers can be found in the Appendix \ref{appendix:loss}. In fact, other layer classifiers exhibit similar trends.}, we observe that both classifiers show similar trends in biased training instances. By contrast, the low-layer classifier possesses higher training loss than the high-layer one on anti-biased training instances. 
\begin{figure}[t]
\centering
\includegraphics[width=0.42\textwidth]{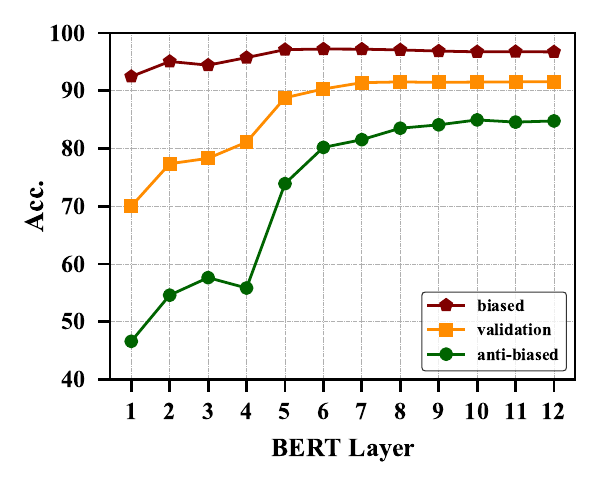}
\vspace{-0.2cm}
\caption{
The prediction performance of layer-specific classifiers on the biased validation instances, the validation set, and the anti-biased validation instances. On the biased validation instances, low-layer classifiers $f_l$ ($1$$\leq$$l$$\leq$$5$) perform slightly worse than high-layer ones, but much worse on anti-biased validation instances.}

\label{fig:bias_learning}
\end{figure}
Next, we compare the accuracies of classifiers on different validation instances. From Figure \ref{fig:bias_learning}, we can find that all classifiers exhibit similar performance on biased validation instances and suffer from performance degradation on anti-biased validation instances. Meanwhile, low-layer classifiers $f_l$ ($1$$\leq$$l$$\leq$$5$) perform worse than high-layer ones on anti-biased validation instances. 

Based on the above experimental results, we can draw the following conclusions: 1) The low- and high-layer sentence representations encode common biased features, which explains that low- and high-layer classifiers show similar loss trends on biased training instances and perform almost well on biased validation instances; 2) The low-layer sentence representations encode less useful task-relevant unbiased features than the high-layer ones so that their classifiers have higher losses on anti-biased training instances and obtain worse results on anti-biased validation instances.

\section{Self-debiasing Framework for Transformer Models}
\label{sec:debiasing-framework}
\begin{figure*}[!th]
\centering
\includegraphics[width=0.8\textwidth]{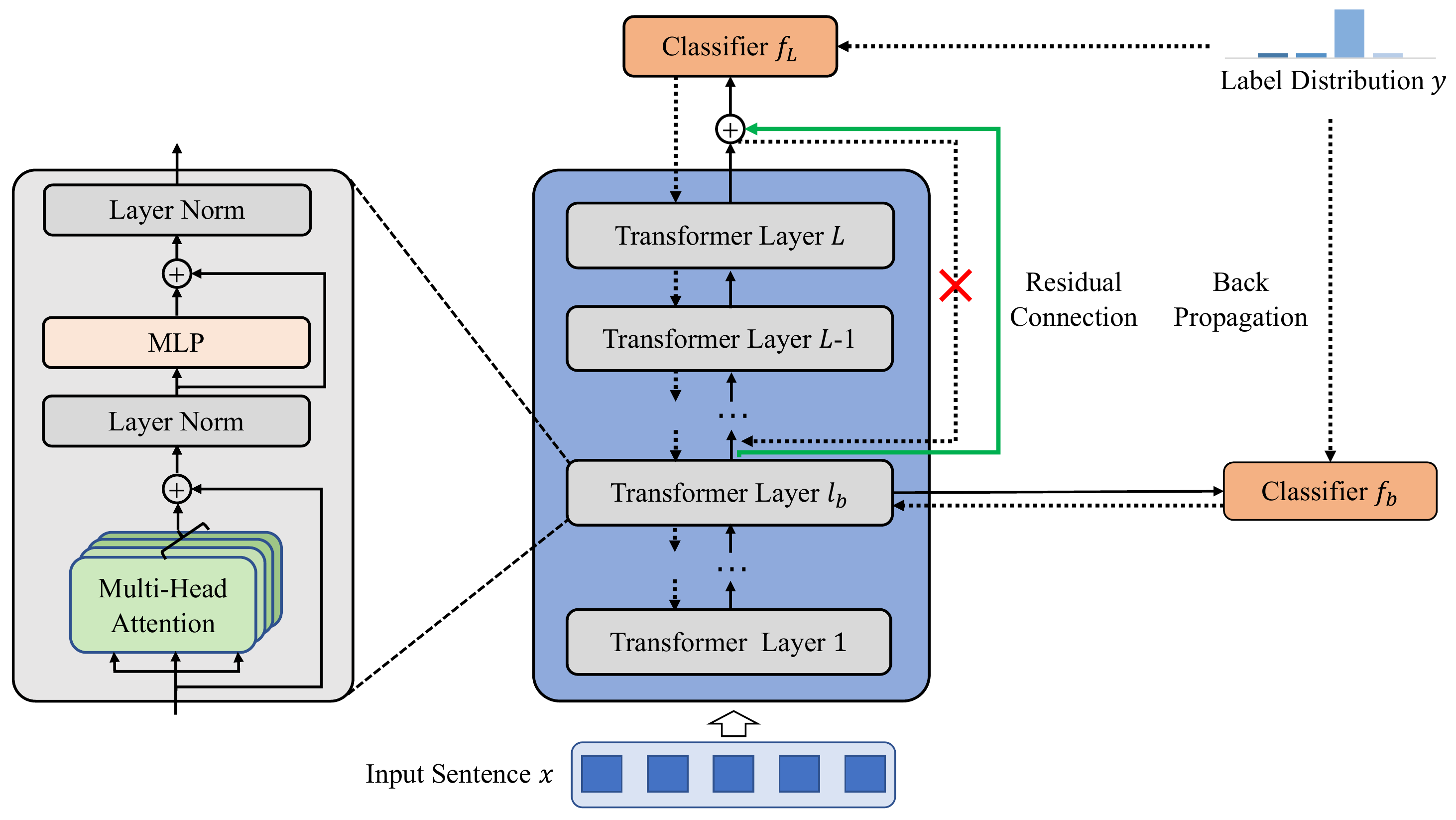}
\caption{
Overview of our framework. In addition to the top-layer classifier $f_L$ based on the top-layer sentence representation $h_{\texttt{CLS}}^L$,
we select a low layer $l_b$ and stack a classifier $f_b$ on its sentence representation $h_{\texttt{CLS}}^{l_b}$. 
Then, we introduce a residual connection feeding the sum of $h_{\texttt{CLS}}^{l_b}$ and $h_{\texttt{CLS}}^{L}$ to $f_L$. 
Through the model training, $h_{\texttt{CLS}}^{l_b}$ will encode common biased features, which have also been encoded in $h_{\texttt{CLS}}^{L}$, and thus $h_{\texttt{CLS}}^{L}$ is encouraged to focus on unbiased features. Notably, we turn off the gradient calculation of the residual connection to avoid the influence of the $f_L$ loss on the representation learning of $h_{\texttt{CLS}}^{l_b}$. The green line denotes our introduced residual connection, and dash lines denote the backpropagation process.
}
\label{fig:framework}
\end{figure*}
Based on the above analyses, we propose a framework for Transformer models by employing a residual connection to exploit the low-layer sentence representation to debias the top-layer sentence representation. Generally, it involves the following two steps.

\paragraph{Step1: Low-layer Sentence Representation Learning.}
\label{sec:bias-features}

As shown in Figure \ref{fig:framework}, given an input sentence $x$, we first employ a Transformer encoder to obtain the contextual representation for each token. Then, we select a low layer $l_b$ and stack a classifier $f_b$ on its sentence representation $h_{\texttt{CLS}}^{l_b}$, which we directly use the contextual representation of \texttt{[CLS]}. As analyzed in Section \ref{ssec:bias-prove-exp}, $h_{\texttt{CLS}}^{l_b}$ will encode common biased features that would also be encoded by top-layer sentence representation $h_{\texttt{CLS}}^{L}$ as the training of $f_b$ goes on. Finally, we obtain the probability distribution $p_b$ over labels as follows:

\begin{equation}
    p_b = \operatorname{Softmax}(W_bh_{\texttt{CLS}}^{l_b}+b_b),
\end{equation}
where $W_b$ and $b_b$ are the learnable parameters. Here, we train the classifier $f_b$ using the commonly-used cross-entropy loss:
\begin{equation}
    \mathcal{L}_b=-\sum_{i=1}^Ky^{(i)}\cdot \operatorname{log}(p_b^{(i)}),
\end{equation}
where $K$ denotes the number of labels, $y^{(i)}$ equals to 1 if the $i$-th label is the golden label, and 0 for other labels.

\paragraph{Step2: Debiasing with a Residual Connection.}
We introduce a residual connection \cite{he2016deep} into our framework, which allows us to exploit the low-layer sentence representation $h_{\texttt{CLS}}^{l_b}$ to debias the top-layer one $h_{\texttt{CLS}}^L$.

Specifically, through this residual connection, we use the sum of $h_{\texttt{CLS}}^{l_b}$ and $h_{\texttt{CLS}}^L$ as the input of the top-layer classifier $f_L$ instead of $h_{\texttt{CLS}}^L$. Formally, the probability distribution output by $f_L$ is calculated as follows:
\begin{equation}
    p_L = \operatorname{Softmax}(W_L(h_{\texttt{CLS}}^{l_b}+h_{\texttt{CLS}}^L)+b_L),
    \label{eq:top_layer_classification}
\end{equation}
where $W_L$ and $b_L$ are also trainable parameters. Note that $f_L$ has the same architecture but different parameters with $f_b$, supervised by a cross-entropy loss:
\begin{equation}
    \mathcal{L}_L=-\sum_{i=1}^Ky^{(i)}\cdot \operatorname{log}(p_L^{(i)}).
\end{equation}

The effectiveness of our design may be attributed to two factors: 1) As stated in Section 3.2, the low-layer representation contains fewer unbiased features than the high-layer ones. This indicates that $h_{\texttt{CLS}}^L$ will encode the unbiased features that are not encoded in $h_{\texttt{CLS}}^{l_b}$. 
2) As $h_{\texttt{CLS}}^{l_b}$ already encodes the common bias features, $h_{\texttt{CLS}}^L$ is encouraged to ignore the common biased features which already encoded in $h_{\texttt{CLS}}^{l_b}$.

Finally, the whole training objective is defined as follows:
\begin{equation}
    \mathcal{L} = \mathcal{L}_b + \mathcal{L}_L.
\end{equation}
Please notice that during training, we turn off the gradient calculation of the residual connection to remove the effect of $\mathcal{L}_L$ on the learning of $h_{\texttt{CLS}}^{l_b}$. During inference, we remove $h_{\texttt{CLS}}^{l_b}$ from Equation \ref{eq:top_layer_classification} and directly use $h_{\texttt{CLS}}^L$ to make predictions\footnote{Further discussion on the model inference can be found in the Appendix \ref{appendix:infer}.}.

\section{Experiments}
\label{sec:experiments}

\subsection{Setup}
\label{ssec:experiment-setup}

\paragraph{Tasks and Datasets}
We conduct several groups of experiments on three common NLU tasks: natural language inference, fact verification, and paraphrase identification. The datasets of each task contain a training set, a validation set, and its corresponding OOD test set. 

\begin{itemize}
\item \textbf{Natural Language Inference (NLI)} is to predict the entailment relationship between the pair of premise and hypothesis. We use the MNLI dataset as the ID set and train various NLI models on it \cite{ DBLP:conf/naacl/WilliamsNB18}, and evaluate them on the OOD test set---HANS \cite{hans2019}. HANS is designed to test whether NLI models make predictions based on three fallible syntactic heuristics: lexical overlap heuristic, subsequence heuristic, and constituent heuristic.
\vspace{-0.2cm}
\item \textbf{Fact Verification (FactVer)} aims to identify whether a claim is supported or refuted by the given evidence text. We adopt the FEVER dataset \cite{thorne2018fact} as the ID set to train models, and assess the model abilities on the OOD test set---FeverSymmetric (Symm.) \cite{feversymm2019}, which is created to reduce claim-only biases.
\vspace{-0.2cm}
\item \textbf{Paraphrase Identification (ParaIden)} is to predict whether the given question pair is duplicate or non-duplicate in semantics. We use the QQP dataset as the ID set to train models, and evaluate model performance on the OOD test set---PAWS \cite{zhang-etal-2019-paws}, which investigates whether the model exploits word overlapping to make predictions.
The basic statistics of all datasets used in our experiments are shown in Table \ref{tab:data_statistics}.
\end{itemize}

\begin{table}[tb]
\small
\center
\setlength{\tabcolsep}{4.5mm}{
\begin{tabular}{l|c c|c}
\toprule[1pt]
\multirow{2}{*}{\textbf{Task}} & \multicolumn{2}{c|}{\textbf{ID Set}} & \multirow{2}{*}{\makecell[c]{\textbf{OOD} \\ \textbf{Test Set}}} \\
\cline{2-3} 
 & Train & Val & \\
\hline
\textbf{NLI}     & 392K  & 19K & 30K             \\
\textbf{ParaIden}      & 363K  & 40K & 8K              \\
\textbf{FactVer}   & 242K  & 16K & 0.7K            \\ 
\bottomrule[1pt]
\end{tabular}
}
\caption{The basic statistics of datasets for three NLU tasks, including the ID Set and the OOD Test Set, where Val refers to the validation set.}
\label{tab:data_statistics}
\end{table}

\paragraph{Baselines}
Most previous debiasing methods involve two stages: biased instance identification and debiasing models. We select several popular methods for each stage and compare their combinations with our framework.

Here, our baseline methods for biased instance identification include:
\begin{itemize}
\item \textbf{Known-Bias} \cite{gururangan-etal-2018-annotation, utama-etal-2020-mind, du-etal-2021-towards}. These approaches quantify the bias degree of each training instance via data statistics or researchers' insights. Then the instances with high bias degree are regarded as biased ones and used to train a bias-only model. 
\vspace{-0.15cm}
\item \textbf{Self-debias}
\cite{utama-etal-2020-towards}.
These approaches train a bias-only model based on partial training data to identify biased instances automatically. 
\end{itemize}

Besides, we select three widely-used debiasing methods for comparison. 
\begin{itemize}
\item \textbf{Re-weighting (RW)} \cite{clark-etal-2019-dont}. This method aims to reduce the contribution of each biased instance on the overall training loss by assigning it with a scalar weight. 
\vspace{-0.15cm}
\item \textbf{Product-of-expert (PoE)} \cite{clark-etal-2019-dont}. It trains the main model in an ensemble manner with the bias-only model, which is trained in advance and uses biased features to make predictions. By doing so, the main model is encouraged to focus on unbiased features and thus becomes more robust.
\vspace{-0.15cm}
\item \textbf{Confidence Regularization (CR)} \cite{utama-etal-2020-mind}. It trains a bias-only model and a teacher model. The output probability distribution of the latter is adjusted with that of the former. Then the re-scaled output distribution is used to enhance a main model.
\end{itemize}

Finally, we also compare our framework with an \textbf{End2End} framework \cite{ghaddar-etal-2021-end}. In this framework, a shallow model and a main model are simultaneously but respectively trained based on the low-layer and the top-layer sentence representations, during which these two models interchangeably re-weight the importance of instances.

To facilitate the subsequent descriptions, we name our framework as \textbf{DeRC}. Besides, we report the performance of a variant of our framework: \textbf{DePoE}. In this variant, we first identify biased training instances according to the low-layer output probabilities and then apply the PoE method to debias the model.

\paragraph{Our Implementations}
\label{details}
To ensure fair comparison, we use \texttt{BERT-base} to develop DeRC\footnote{We also use \texttt{RoBERTa-base} to develop DeRC, proving that DeRC is still effective on other Transformer-based models. Details can be found in Appendix
\ref{appendix:roberta}.}. During the process of fine-tuning models on each dataset, we follow the standard setup \cite{devlin-etal-2019-bert} to construct inputs and use the hidden state of token \texttt{[CLS]} for classification. For each task, we use a batch size of 32 and fine-tune the model for 5 epochs with the learning rate 5e-5. Besides, we select Adam \cite{DBLP:journals/corr/KingmaB14} as the optimizer to update parameters. 

We evaluate the model performance on the validation sets and the corresponding OOD test sets. Following \citet{utama-etal-2020-towards}, we use accuracy (Acc.) as the main metric for three tasks. 
In addition, we evaluate the interpretability results on the QQP dataset using the metrics proposed by \citet{wang2022fine}. Please see Section \ref{ssec:result-analysis} for details.

\defcitealias{utama-etal-2020-towards}{(Utama et al. 2020b)}
\defcitealias{clark-etal-2019-dont}{(Clark et al. 2019)}
\defcitealias{utama-etal-2020-mind}{(Utama et al. 2020a)}

\begin{table*}[!t]
\centering
\resizebox{\textwidth}{!}{
\renewcommand{\arraystretch}{1.}
\begin{tabular}{l|ccr|ccr|ccr}
\toprule[1pt]
\multirow{2}{*}{\textbf{Model}} & \multicolumn{3}{c|}{\textbf{MNLI}}     & \multicolumn{3}{c|}{\textbf{FEVER}}     & \multicolumn{3}{c}{\textbf{QQP}}         \\
                                  & \textbf{Val} & \textbf{HANS} & \textbf{$\Delta$} & \textbf{Val} & \textbf{Symm.} & \textbf{$\Delta$} & \textbf{Val} & \textbf{PAWS} & \textbf{$\Delta$} \\ \hline
BERT                     & \textbf{84.5}         & 62.3          & -              & 85.9         & 64.4           & -              & \textbf{91.0}         & 33.5          & -              \\
\hline
Known-Bias + RW \citetalias{clark-etal-2019-dont}   & 83.5  & 69.2  & +6.9           & 84.6         & 66.5          & +2.1           & -        & -         & -           \\
Known-Bias + PoE \citetalias{clark-etal-2019-dont}   & 82.9  & 67.9  & +5.6           & 86.4         & 69.1           & +4.7           & -         & -          & -           \\
Known-Bias + CR \citetalias{utama-etal-2020-mind}   & \textbf{84.5}         &  69.1          & +6.8           & 86.4         & 66.2           & +1.6           & 89.1         & 40.0          & +6.5           \\
Self-debias + RW \citetalias{utama-etal-2020-towards} & 82.3 & 69.7 & +7.4 & 87.1 & 65.5 & +1.1 & 85.2 & 57.4 & +23.9 \\
Self-debias + PoE \citetalias{utama-etal-2020-towards}  & 81.9 & 66.8 & +4.5 & 85.9 & 65.8 & +1.4 & - & - & - \\
Self-debias + CR \citetalias{utama-etal-2020-towards} & 84.3 & 67.1 & +4.8 & 87.5 & 66.0 & +1.6 & 89.0 & 43.0 & +9.5 \\
\hline
End2End \cite{ghaddar-etal-2021-end}                       & 83.2         & 71.2          & +8.9           & 86.9         & -           & -           & 90.2            & 46.5             & +13.0              \\ 
\hline
DePoE                    & 83.6            & 62.6             & +0.3              & 78.0            & 68.0              & +3.6              & 79.7            & 59.2             & +25.7              \\
DeRC                     & 82.8         & \textbf{72.6}          & \textbf{+10.3}          & \textbf{88.1}         & \textbf{71.9}           & \textbf{+7.5}          & 88.4         & \textbf{59.8}          & \textbf{+26.3}          \\

\bottomrule[1pt]
\end{tabular}
}
\caption{Experimental results on two sets: 1) the validation sets of MNLI, FEVER, QQP; 2) their corresponding OOD test sets. The results for QQP are directly cited from \cite{ghaddar-etal-2021-end} and the other results are cited from the corresponding papers.
Values of \textbf{$\Delta$} denote the performance gaps between debiasing methods and BERT on the OOD test sets.
}
\label{tab:main_results}
\end{table*}
\begin{figure*}[!ht]
\centering
    \subfigure[NLI]{   
    \includegraphics[width=0.31\linewidth]{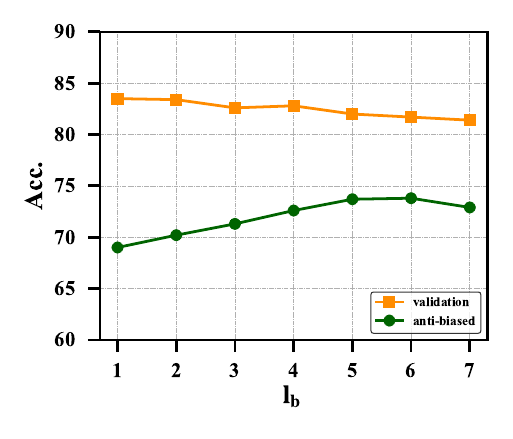}
    }
    \subfigure[FactVer]{ 
    \includegraphics[width=0.31\linewidth]{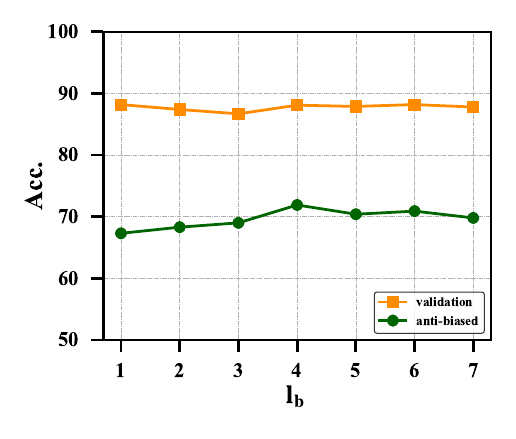}
    }
    \subfigure[ParaIden]{
    \includegraphics[width=0.312\linewidth]{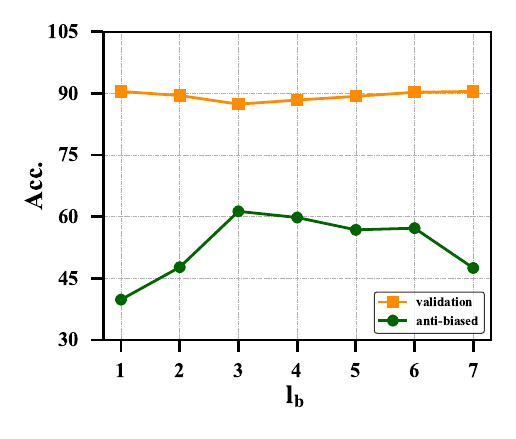}
    }
   
    \vspace{-0.35cm}
	\caption{The
		performance of DeRC with different selected low layer $l_b$. Green lines denote the results of the validation sets, and orange ones denote those of the OOD test sets.
	}
	\label{fig:various_layers}
\end{figure*}
\subsection{Effect of the Chosen Low Layer \textbf{$l_b$}} 
Under our framework, the chosen low layer $l_b$ is a crucial hyperparameter for biased sentence representation learning. Based on our analysis in Section \ref{ssec:bias-prove-exp}, we argue that the performance gap of the low-layer classifier between the biased and anti-biased instances should be as large as possible. In this way, the classifier of low-layer $l_b$ would encode sufficient biased features while encoding less task-relevant useful unbiased features. However, a too-small value of $l_b$ is not an ideal choice, since such a low-layer classifier will not comprehensively capture biased features. 
As demonstrated in \cite{clark-etal-2019-dont}, 
the lowest layers, like the 1st and 2nd layers, would ignore syntactic features, which may also belong to biased features. Therefore, we need to choose a layer whose representation may contain more potentially biased features. Finally, taking the result of Figure \ref{fig:bias_learning} into account as well, we select $l_b$ as 4.

To further examine the impact of $l_b$, we vary its value from 1 to 7 and evaluate the performance of DeRC on the validation and the OOD test sets. As shown in Figure \ref{fig:various_layers}, DeRC consistently performs well on all OOD test sets for all $l_b$ , indicating that $l_b$ is task-agnostic and generic for all datasets. Furthermore, we find that the model's performance deteriorates slightly when $l_b$ is within the range of 1 to 2, supporting our belief that $l_b$ should not be too small. 

\subsection{Main Results}
\label{ssec:main-results}
Experimental results are reported in Table \ref{tab:main_results}\footnote{Some results from previous works are missing because they did not report. We have not reproduced these results due to the absence of sufficient detail for reproduction in their respective papers.}. We can observe that DeRC achieves the best performance on all OOD test sets of three tasks, setting a new SOTA. In particular, on the Symm. set, DeRC improves accuracy by 7.5\% than BERT, while the previous best model (Known-Bias+PoE) only brings a gain of 4.7\%. Besides, DeRC reaches the best performance on PAWS, surpassing the previous best work (Known-Bias+CR), while bringing a much less performance drop on QQP (2.6\% v.s. 5.8\%). In addition, DeRC achieves the most significant accuracy improvement on FEVER. Thus, we confirm that DeRC is effective in improving the model performance on OOD test sets and harmless on the validation sets.

Furthermore, the comparison between DeRC and its variant DePoE also proves the effectiveness of the residual connection. Unlike DeRC uses a residual connection, DePoE uses the probabilities output by the low-layer classifier to debias models. The results show that the residual connection enables DeRC to achieve a better trade-off between the performance drop on the validation sets and the improvement on the OOD test sets.

\begin{figure}[!t]
\centering
\includegraphics[width=0.4\textwidth]{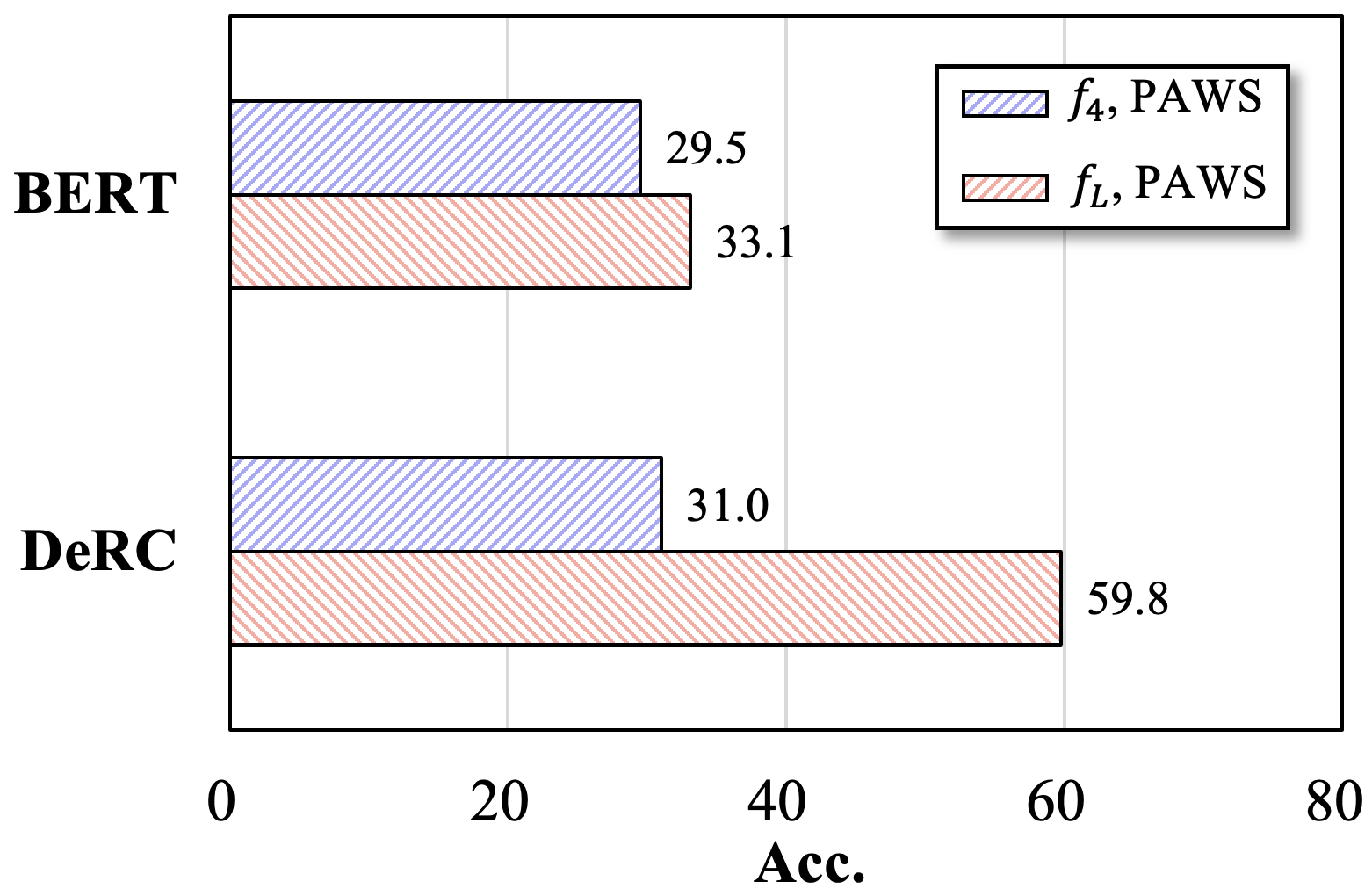}
\caption{
Performance on PAWS of the 4th-layer and top-layer classifiers in BERT and DeRC.
}
\label{fig:impact_of_RC}
\end{figure}

\subsection{Analysis}
\label{ssec:result-analysis}
Moreover, we conduct more analyses to verify the effectiveness of DeRC.

\paragraph{Impact of Residual Connection}
In this experiment, we use QQP to train BERT and DeRC, and compare their prediction accuracies on PAWS. Similar to DeRC, we stack two classifiers on the two layers of BERT: one is the top-layer classifier, and the other is the 4th-layer classifier. As shown in Figure \ref{fig:impact_of_RC}, the accuracy gap between the 4th-layer and top-layer classifiers of DeRC is more significant than that of BERT. Note that BERT and DeRC are similar in architecture, and the only difference is that DeRC introduces a residual connection to debias the top-layer sentence representation.
Thus, we confirm that the residual connection significantly improves the model generalizability.

\paragraph{Interpretability evaluation} In the field of post-hoc interpretation research, many studies intend to interpret the model prediction by assigning each input token with an importance score, which quantifies its impact on the prediction \cite{simonyan2014deep, smilkov2017smoothgrad, jain-wallace-2019-attention}. In this way, the most important tokens can form the rationale supporting the prediction. Inspired by these studies, we use QQP to train DeRC and then report interpretability results on the validation set released by \citet{wang2022fine}, which provides annotated rationales and corresponding evaluation metrics for interpretability. 

Concretely, we adopt the attention-based interpretation method \cite{jain-wallace-2019-attention} to assign input tokens with importance scores and then follow \citet{wang2022fine} to select the top-$k$ important tokens as the rationale. Afterwards, as implemented in \cite{deyoung-etal-2020-eraser, wang2022fine}, we use four metrics to evaluate the model interpretability from the perspective of plausibility and faithfulness: 

\begin{table}[t]
\resizebox{\linewidth}{!}{
\renewcommand{\arraystretch}{1.3}
\centering
\small
\scalebox{0.98}{
\begin{tabular}{l|c|c|ccc}
\toprule[1pt]
\multirow{2}{*}{\textbf{Models}} & \multirow{2}{*}{\textbf{Acc.}} & \textbf{Plausibility} & \multicolumn{3}{c}{\textbf{Faithfulness}}\\
\cline{3-6}
 & & \textbf{Token F1} & \textbf{MAP} & \textbf{Suff.} $\downarrow$ & \textbf{Comp.} \\
 \hline
BERT & 90.07\% & 58.31\% & 71.24\%  & 0.1531 & 0.3217 \\ 
DeRC & \makecell[c]{\textbf{91.13\%}  } & \makecell[c]{\textbf{62.25\%} }   & \makecell[c]{\textbf{75.62\%} } & \makecell[c]{\textbf{0.0922}  }  & \makecell[c]{\textbf{0.3843}  } \\ 
\bottomrule[1pt]
\end{tabular}}
}
\caption{Evaluation results of interpretability. The metric with $\downarrow$ means the lower the score is, the better the performance achieves. For all other metrics, a high score represents good performance.
}
\label{tab:interpretability_results}

\end{table}

\begin{itemize}
    \item \textbf{Token-F1}. It is used to evaluate plausibility by measuring the token overlap between the model-generated and human-annotated rationales. The higher the Token-F1 is, the more plausible the rationale is.
    \item \textbf{MAP}. This metric measures the consistency of rationales under perturbations, and is used to evaluate faithfulness. A high MAP represents high faithfulness.
    \item \textbf{Sufficiency} (Suff.) and \textbf{Comprehensiveness} (Comp.). Both two metrics are used to assess the degree of the provided rationale reflecting the prediction. 
    A faithful rationale should have a low sufficiency score and a high comprehensiveness score.
\end{itemize}

From Table \ref{tab:interpretability_results}, we can find that DeRC outperforms BERT on all metrics, that is, the rationales provided by DeRC are more plausible and faithful. Thus, we confirm that DeRC can improve the model ability of understanding.

\section{Conclusions}
In this work, we have proposed DeRC for Transformer-based NLU models, which utilizes the biased sentence representation learned by the low-layer classifier to debias the top-layer sentence representation. Compared with previous studies, DeRC is more efficient as it does not require manual analysis or the use of an additional bias-only model. We conduct extensive experiments on commonly-used datasets of three NLU tasks. Experimental results show that DeRC can achieve better performance on OOD test sets, while maintaining comparable performance on validation sets. In addition, DeRC can improve the ability of understanding. 

In the future, we will continue to explore the low-layer representations for better performance trade-off between the validation and OOD test sets during inference. In addition, we plan to apply DeRC to other NLU tasks, such as sentiment analysis, machine reading comprehension, and so on. Finally, we will study whether DeRC is suitable for natural language generation tasks.

\section*{Limitations}
The limitations of this work are the following aspects: 1) The proposed DeRC is only applicable to models based on the Transformer architecture; 2) We only focus on the result of NLU tasks, ignoring further discussions on its contribution to other tasks.

\section*{Ethical Statements}
This paper proposes a self-debiasing framework for Transformer-based models. Typically, our proposed framework utilizes the low-layer sentence representation to debias the top-layer one, which improves the model robustness to spurious correlation and the ability of understanding. This study will not pose ethical issues. All the datasets used in this paper are publicly available and widely adopted by researchers to test the performance of debiasing frameworks. Besides, this paper does not involve any data collection and release, and thus there exist no privacy issue.

\bibliographystyle{acl_natbib}
\bibliography{custom}

\begin{thebibliography}{43}
\expandafter\ifx\csname natexlab\endcsname\relax\def\natexlab#1{#1}\fi

\bibitem[{Belinkov et~al.(2019)Belinkov, Poliak, Shieber, Van~Durme, and
  Rush}]{belinkov-etal-2019-adversarial}
Yonatan Belinkov, Adam Poliak, Stuart Shieber, Benjamin Van~Durme, and
  Alexander Rush. 2019.
\newblock On adversarial removal of hypothesis-only bias in natural language
  inference.
\newblock In \emph{SEM 2019}.

\bibitem[{Clark et~al.(2019)Clark, Yatskar, and
  Zettlemoyer}]{clark-etal-2019-dont}
Christopher Clark, Mark Yatskar, and Luke Zettlemoyer. 2019.
\newblock Don{'}t take the easy way out: Ensemble based methods for avoiding
  known dataset biases.
\newblock In \emph{EMNLP 2019}.

\bibitem[{Clark et~al.(2020)Clark, Yatskar, and
  Zettlemoyer}]{clark-etal-2020-learning}
Christopher Clark, Mark Yatskar, and Luke Zettlemoyer. 2020.
\newblock Learning to model and ignore dataset bias with mixed capacity
  ensembles.
\newblock In \emph{Findings of EMNLP 2020}.

\bibitem[{Devlin et~al.(2019)Devlin, Chang, Lee, and
  Toutanova}]{devlin-etal-2019-bert}
Jacob Devlin, Ming-Wei Chang, Kenton Lee, and Kristina Toutanova. 2019.
\newblock {BERT}: Pre-training of deep bidirectional transformers for language
  understanding.
\newblock In \emph{NAACL 2019}.

\bibitem[{DeYoung et~al.(2020)DeYoung, Jain, Rajani, Lehman, Xiong, Socher, and
  Wallace}]{deyoung-etal-2020-eraser}
Jay DeYoung, Sarthak Jain, Nazneen~Fatema Rajani, Eric Lehman, Caiming Xiong,
  Richard Socher, and Byron~C. Wallace. 2020.
\newblock {ERASER}: {A} benchmark to evaluate rationalized {NLP} models.
\newblock In \emph{ACL 2020}.

\bibitem[{Du et~al.(2021)Du, Manjunatha, Jain, Deshpande, Dernoncourt, Gu, Sun,
  and Hu}]{du-etal-2021-towards}
Mengnan Du, Varun Manjunatha, Rajiv Jain, Ruchi Deshpande, Franck Dernoncourt,
  Jiuxiang Gu, Tong Sun, and Xia Hu. 2021.
\newblock Towards interpreting and mitigating shortcut learning behavior of
  {NLU} models.
\newblock In \emph{NAACL 2021}.

\bibitem[{Ganin and Lempitsky(2015)}]{ganin2015unsupervised}
Y~Ganin and V~Lempitsky. 2015.
\newblock Unsupervised domain adaptation by backpropagation.
\newblock In \emph{ICML 2015}.

\bibitem[{Geirhos et~al.(2020)Geirhos, Jacobsen, Michaelis, Zemel, Brendel,
  Bethge, and Wichmann}]{geirhos2020shortcut}
Robert Geirhos, J{\"o}rn-Henrik Jacobsen, Claudio Michaelis, Richard Zemel,
  Wieland Brendel, Matthias Bethge, and Felix~A Wichmann. 2020.
\newblock Shortcut learning in deep neural networks.
\newblock \emph{Nature Machine Intelligence}.

\bibitem[{Ghaddar et~al.(2021)Ghaddar, Langlais, Rezagholizadeh, and
  Rashid}]{ghaddar-etal-2021-end}
Abbas Ghaddar, Phillippe Langlais, Mehdi Rezagholizadeh, and Ahmad Rashid.
  2021.
\newblock End-to-end self-debiasing framework for robust {NLU} training.
\newblock In \emph{Findings of ACL 2021}.

\bibitem[{Gururangan et~al.(2018)Gururangan, Swayamdipta, Levy, Schwartz,
  Bowman, and Smith}]{gururangan-etal-2018-annotation}
Suchin Gururangan, Swabha Swayamdipta, Omer Levy, Roy Schwartz, Samuel Bowman,
  and Noah~A. Smith. 2018.
\newblock Annotation artifacts in natural language inference data.
\newblock In \emph{NAACL 2018}.

\bibitem[{He et~al.(2019)He, Zha, and Wang}]{he-etal-2019-unlearn}
He~He, Sheng Zha, and Haohan Wang. 2019.
\newblock Unlearn dataset bias in natural language inference by fitting the
  residual.
\newblock In \emph{Workshop of DeepLo 2019}.

\bibitem[{He et~al.(2016)He, Zhang, Ren, and Sun}]{he2016deep}
Kaiming He, Xiangyu Zhang, Shaoqing Ren, and Jian Sun. 2016.
\newblock Deep residual learning for image recognition.
\newblock In \emph{CVPR 2016}.

\bibitem[{Hewitt and Manning(2019)}]{hewitt-manning-2019-structural}
John Hewitt and Christopher~D. Manning. 2019.
\newblock {A} structural probe for finding syntax in word representations.
\newblock In \emph{NAACL 2019}.

\bibitem[{Jain and Wallace(2019)}]{jain-wallace-2019-attention}
Sarthak Jain and Byron~C. Wallace. 2019.
\newblock {A}ttention is not {E}xplanation.
\newblock In \emph{NAACL 2019}.

\bibitem[{Jawahar et~al.(2019)Jawahar, Sagot, and Seddah}]{bert-structure2019}
Ganesh Jawahar, Beno{\^\i}t Sagot, and Djam{\'e} Seddah. 2019.
\newblock What does bert learn about the structure of language?
\newblock In \emph{ACL 2019}.

\bibitem[{Karimi~Mahabadi et~al.(2020)Karimi~Mahabadi, Belinkov, and
  Henderson}]{karimi-mahabadi-etal-2020-end}
Rabeeh Karimi~Mahabadi, Yonatan Belinkov, and James Henderson. 2020.
\newblock End-to-end bias mitigation by modelling biases in corpora.
\newblock In \emph{ACL 2020}.

\bibitem[{Kingma and Ba(2015)}]{DBLP:journals/corr/KingmaB14}
Diederik~P. Kingma and Jimmy Ba. 2015.
\newblock Adam: {A} method for stochastic optimization.
\newblock In \emph{ICLR 2015}.

\bibitem[{Lee et~al.(2021)Lee, Won, Kim, Lee, Park, and Jung}]{lee2021crossaug}
Minwoo Lee, Seungpil Won, Juae Kim, Hwanhee Lee, Cheoneum Park, and Kyomin
  Jung. 2021.
\newblock Crossaug: A contrastive data augmentation method for debiasing fact
  verification models.
\newblock \emph{Information and Knowledge Management}.

\bibitem[{Liu et~al.(2019)Liu, Ott, Goyal, Du, Joshi, Chen, Levy, Lewis,
  Zettlemoyer, and Stoyanov}]{liu2019roberta}
Yinhan Liu, Myle Ott, Naman Goyal, Jingfei Du, Mandar Joshi, Danqi Chen, Omer
  Levy, Mike Lewis, Luke Zettlemoyer, and Veselin Stoyanov. 2019.
\newblock Roberta: A robustly optimized bert pretraining approach.
\newblock \emph{arXiv preprint arXiv:1907.11692}.

\bibitem[{Lyu et~al.(2023)Lyu, Li, Yang, de~Rijke, Ren, Zhao, Yin, and
  Ren}]{lyu2022feature}
Yougang Lyu, Piji Li, Yechang Yang, Maarten de~Rijke, Pengjie Ren, Yukun Zhao,
  Dawei Yin, and Zhaochun Ren. 2023.
\newblock Feature-level debiased natural language understanding.
\newblock In \emph{AAAI 2023}.

\bibitem[{McCoy et~al.(2019)McCoy, Pavlick, and Linzen}]{hans2019}
Tom McCoy, Ellie Pavlick, and Tal Linzen. 2019.
\newblock Right for the wrong reasons: Diagnosing syntactic heuristics in
  natural language inference.
\newblock In \emph{ACL 2019}.

\bibitem[{Nie et~al.(2020)Nie, Williams, Dinan, Bansal, Weston, and
  Kiela}]{nie2020adversarial}
Yixin Nie, Adina Williams, Emily Dinan, Mohit Bansal, Jason Weston, and Douwe
  Kiela. 2020.
\newblock Adversarial nli: A new benchmark for natural language understanding.
\newblock In \emph{ACL 2020}.

\bibitem[{Poliak et~al.(2018)Poliak, Naradowsky, Haldar, Rudinger, and
  Van~Durme}]{poliak-etal-2018-hypothesis}
Adam Poliak, Jason Naradowsky, Aparajita Haldar, Rachel Rudinger, and Benjamin
  Van~Durme. 2018.
\newblock Hypothesis only baselines in natural language inference.
\newblock In \emph{SEM 2018}.

\bibitem[{Sanh et~al.(2020)Sanh, Wolf, Belinkov, and Rush}]{sanh2020learning}
Victor Sanh, Thomas Wolf, Yonatan Belinkov, and Alexander~M Rush. 2020.
\newblock Learning from others' mistakes: Avoiding dataset biases without
  modeling them.
\newblock \emph{arXiv preprint arXiv:2012.01300}.

\bibitem[{Schuster et~al.(2019)Schuster, Shah, Yeo, Roberto Filizzola~Ortiz,
  Santus, and Barzilay}]{feversymm2019}
Tal Schuster, Darsh Shah, Yun Jie~Serene Yeo, Daniel Roberto Filizzola~Ortiz,
  Enrico Santus, and Regina Barzilay. 2019.
\newblock Towards debiasing fact verification models.
\newblock In \emph{EMNLP 2019}.

\bibitem[{Sharma et~al.(2018)Sharma, Allen, Bakhshandeh, and
  Mostafazadeh}]{sharma-etal-2018-tackling}
Rishi Sharma, James Allen, Omid Bakhshandeh, and Nasrin Mostafazadeh. 2018.
\newblock Tackling the story ending biases in the story cloze test.
\newblock In \emph{ACL 2018}.

\bibitem[{Simonyan et~al.(2014)Simonyan, Vedaldi, and
  Zisserman}]{simonyan2014deep}
Karen Simonyan, Andrea Vedaldi, and Andrew Zisserman. 2014.
\newblock Deep inside convolutional networks: Visualising image classification
  models and saliency maps.
\newblock In \emph{ICLR 2014}.

\bibitem[{Smilkov et~al.(2017)Smilkov, Thorat, Kim, Vi{\'e}gas, and
  Wattenberg}]{smilkov2017smoothgrad}
Daniel Smilkov, Nikhil Thorat, Been Kim, Fernanda Vi{\'e}gas, and Martin
  Wattenberg. 2017.
\newblock Smoothgrad: removing noise by adding noise.
\newblock \emph{arXiv preprint arXiv:1706.03825}.

\bibitem[{Snow et~al.(2006)Snow, Vanderwende, and
  Menezes}]{snow-etal-2006-effectively}
Rion Snow, Lucy Vanderwende, and Arul Menezes. 2006.
\newblock Effectively using syntax for recognizing false entailment.
\newblock In \emph{NAACL 2006}.

\bibitem[{Stacey et~al.(2020)Stacey, Minervini, Dubossarsky, Riedel, and
  Rockt{\"a}schel}]{stacey-etal-2020-avoiding}
Joe Stacey, Pasquale Minervini, Haim Dubossarsky, Sebastian Riedel, and Tim
  Rockt{\"a}schel. 2020.
\newblock {A}voiding the {H}ypothesis-{O}nly {B}ias in {N}atural {L}anguage
  {I}nference via {E}nsemble {A}dversarial {T}raining.
\newblock In \emph{EMNLP 2020}.

\bibitem[{Tenney et~al.(2019)Tenney, Das, and Pavlick}]{tenney-etal-2019-bert}
Ian Tenney, Dipanjan Das, and Ellie Pavlick. 2019.
\newblock {BERT} rediscovers the classical {NLP} pipeline.
\newblock In \emph{ACL 2019}.

\bibitem[{Thorne et~al.(2018)Thorne, Vlachos, Cocarascu, Christodoulopoulos,
  and Mittal}]{thorne2018fact}
James Thorne, Andreas Vlachos, Oana Cocarascu, Christos Christodoulopoulos, and
  Arpit Mittal. 2018.
\newblock The fact extraction and verification (fever) shared task.
\newblock In \emph{EMNLP 2018}.

\bibitem[{Tsuchiya(2018)}]{tsuchiya-2018-performance}
Masatoshi Tsuchiya. 2018.
\newblock Performance impact caused by hidden bias of training data for
  recognizing textual entailment.
\newblock In \emph{LREC 2018}.

\bibitem[{Tu et~al.(2020)Tu, Lalwani, Gella, and He}]{tu2020empirical}
Lifu Tu, Garima Lalwani, Spandana Gella, and He~He. 2020.
\newblock An empirical study on robustness to spurious correlations using
  pre-trained language models.
\newblock \emph{Transactions of the Association for Computational Linguistics}.

\bibitem[{Utama et~al.(2020{\natexlab{a}})Utama, Moosavi, and
  Gurevych}]{utama-etal-2020-mind}
Prasetya~Ajie Utama, Nafise~Sadat Moosavi, and Iryna Gurevych.
  2020{\natexlab{a}}.
\newblock Mind the trade-off: Debiasing {NLU} models without degrading the
  in-distribution performance.
\newblock In \emph{ACL 2020}.

\bibitem[{Utama et~al.(2020{\natexlab{b}})Utama, Moosavi, and
  Gurevych}]{utama-etal-2020-towards}
Prasetya~Ajie Utama, Nafise~Sadat Moosavi, and Iryna Gurevych.
  2020{\natexlab{b}}.
\newblock Towards debiasing {NLU} models from unknown biases.
\newblock In \emph{EMNLP 2020}.

\bibitem[{Wang et~al.(2018)Wang, Singh, Michael, Hill, Levy, and
  Bowman}]{wang-etal-2018-glue}
Alex Wang, Amanpreet Singh, Julian Michael, Felix Hill, Omer Levy, and Samuel
  Bowman. 2018.
\newblock {GLUE}: A multi-task benchmark and analysis platform for natural
  language understanding.
\newblock In \emph{Workshop of EMNLP 2018}.

\bibitem[{Wang et~al.(2022)Wang, Shen, Peng, Zhang, Xiao, Liu, Tang, Chen, Wu,
  and Wang}]{wang2022fine}
Lijie Wang, Yaozong Shen, Shuyuan Peng, Shuai Zhang, Xinyan Xiao, Hao Liu,
  Hongxuan Tang, Ying Chen, Hua Wu, and Haifeng Wang. 2022.
\newblock A fine-grained interpretability evaluation benchmark for neural nlp.
\newblock \emph{arXiv preprint arXiv:2205.11097}.

\bibitem[{Williams et~al.(2018)Williams, Nangia, and
  Bowman}]{DBLP:conf/naacl/WilliamsNB18}
Adina Williams, Nikita Nangia, and Samuel~R. Bowman. 2018.
\newblock A broad-coverage challenge corpus for sentence understanding through
  inference.
\newblock In \emph{NAACL 2018}.

\bibitem[{Yaghoobzadeh et~al.(2021)Yaghoobzadeh, Mehri, des Combes, Hazen, and
  Sordoni}]{yaghoobzadeh2021increasing}
Yadollah Yaghoobzadeh, Soroush Mehri, Remi~Tachet des Combes, Timothy~J Hazen,
  and Alessandro Sordoni. 2021.
\newblock Increasing robustness to spurious correlations using forgettable
  examples.
\newblock In \emph{ACL 2021}.

\bibitem[{Zellers et~al.(2018)Zellers, Bisk, Schwartz, and
  Choi}]{zellers-etal-2018-swag}
Rowan Zellers, Yonatan Bisk, Roy Schwartz, and Yejin Choi. 2018.
\newblock {SWAG}: A large-scale adversarial dataset for grounded commonsense
  inference.
\newblock In \emph{EMNLP 2018}.

\bibitem[{Zellers et~al.(2019)Zellers, Holtzman, Bisk, Farhadi, and
  Choi}]{zellers-etal-2019-hellaswag}
Rowan Zellers, Ari Holtzman, Yonatan Bisk, Ali Farhadi, and Yejin Choi. 2019.
\newblock {H}ella{S}wag: Can a machine really finish your sentence?
\newblock In \emph{ACL 2019}.

\bibitem[{Zhang et~al.(2019)Zhang, Baldridge, and He}]{zhang-etal-2019-paws}
Yuan Zhang, Jason Baldridge, and Luheng He. 2019.
\newblock {PAWS}: Paraphrase adversaries from word scrambling.
\newblock In \emph{NAACL 2019}.

\end{thebibliography}
\clearpage
\appendix
\section{Training Loss Curves of other Layer-Specific Classifiers}
\label{appendix:loss}
\begin{figure}[t]
\includegraphics[width=0.48\textwidth]{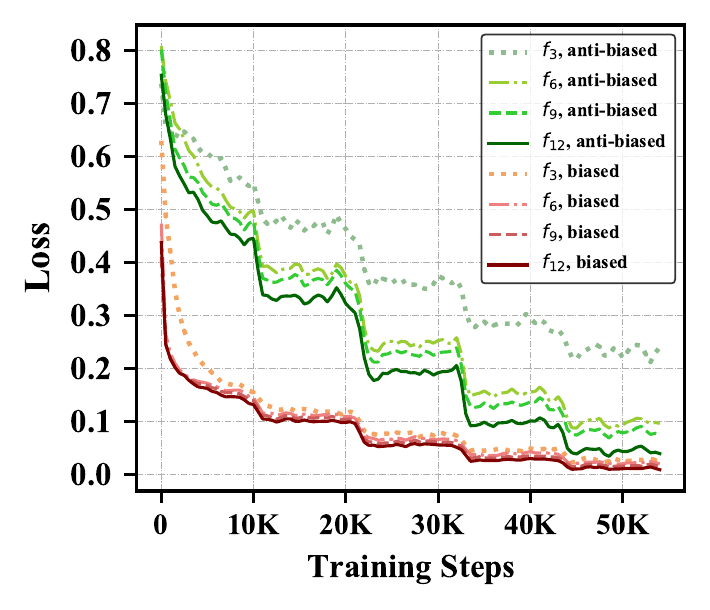}
\vspace{-0.65cm}
\caption{Training loss curves on biased and anti-biased training instances, where $f_i$ represents the $i$-th layer classifier.}
\end{figure}

\section{Further Discussion about the Model Inference}
\label{appendix:infer}
We notice that in addition to the common biased features, the low-layer sentence representation also contains useful unbiased features. Thus, Equation \ref{eq:top_layer_classification} not only encourages the top-layer sentence representation to ignore the common biased features but also makes it discard some useful unbiased features, of which the amount is less than that of biased features, as analyzed in Section \ref{ssec:bias-prove-exp}. To deal with this issue, we reincorporate the low-layer sentence representation into the top-layer classifier in a weighting manner:
\begin{equation}
    p_L = \operatorname{Softmax}(W_L(\alpha*h_{\texttt{CLS}}^{l_b}+(1-\alpha)*h_{\texttt{CLS}}^L)+b_L)
\label{eq:inference}
\end{equation}
where $\alpha$ is used to control the effect of $h_{\texttt{CLS}}^{l_b}$ during inference. 
\begin{figure}[t]
\centering

\includegraphics[width=0.43\textwidth]{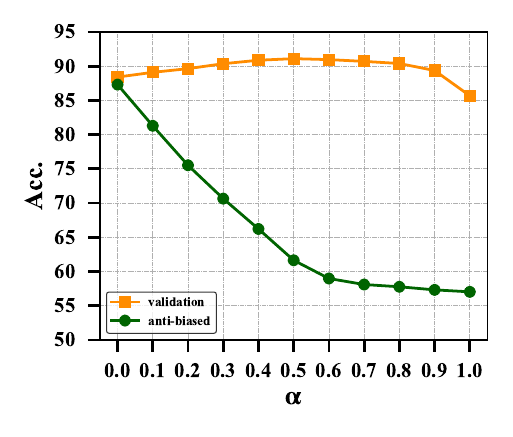}
\vspace{-0.3cm}
\caption{
Performance of models with different $\alpha$ on the MNLI validation set and anti-biased validation instances.
}
\label{fig:debias_lambda}
\end{figure}

  Then, we vary $\alpha$ from 0 to 1 with an interval of 0.1, and compare the model performance on both the validation set and anti-biased validation instances. As shown in Figure \ref{fig:debias_lambda}, although the use of low-layer sentence representation slightly improves the model's performance on the validation set, it significantly degrads the performance on the anti-biased instances as $\alpha$ increases. Therefore, we directly set $\alpha$ to 0 in subsequent experiments. In other words, we will use the top-layer sentence representation for predictions during inference.

\section{Experiments based on RoBERT-base}
\label{appendix:roberta}
\begin{figure}[t]
\centering
\includegraphics[width=0.48\textwidth]{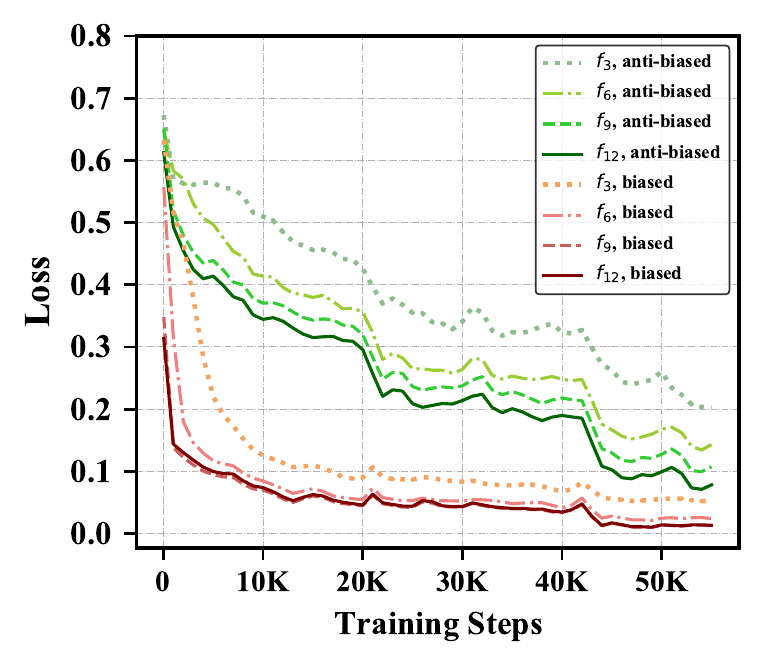}
\vspace{-0.8cm}
\caption{
Training loss curves on biased and anti-biased
training instances, where $f_i$ represents the $i$-th layer
classifier.
}
\label{fig:roberta loss}
\end{figure}

\begin{figure}[t]
\centering
\includegraphics[width=0.48\textwidth]{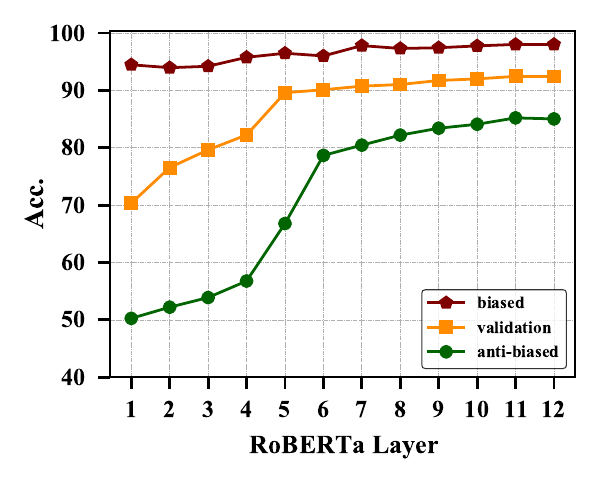}
\vspace{-0.65cm}
\caption{
The prediction performance of layer-specific
classifiers on the validation set, the biased validation
instances, and the anti-biased validation instances.
}
\label{fig:roberta acc}
\end{figure}

As implemented in Section \ref{ssec:bias-prove-exp}, we also conduct a preliminary experiment to analyze the feature learning of different layers of \texttt{RoBERTa-base} \cite{liu2019roberta} from the perspectives of model training and prediction. From Figure \ref{fig:roberta loss} and Figure \ref{fig:roberta acc}, we can find that the training losses and prediction accuracies of the RoBERTa exhibit almost the same trends as those of BERT.

\begin{table*}[!t]
\centering
\renewcommand{\arraystretch}{1.}
\begin{tabular}{l|ccr|ccr|ccr}
\toprule[1pt]
\multirow{2}{*}{\textbf{Model}} & \multicolumn{3}{c|}{\textbf{MNLI}}     & \multicolumn{3}{c|}{\textbf{FEVER}}     & \multicolumn{3}{c}{\textbf{QQP}}         \\
                                  & \textbf{Val} & \textbf{HANS} & \textbf{$\Delta$} & \textbf{Val} & \textbf{Symm.} & \textbf{$\Delta$} & \textbf{Val} & \textbf{PAWS} & \textbf{$\Delta$} \\ \hline
RoBERTa                     & \textbf{87.2}         & 73.5          & -              & \textbf{89.3}        & 66.3           & -              & \textbf{91.5}         & 40.1          & -              \\
\hline
DePoE                    & 84.9            & 75.2             & +1.7             & 87.2            & 69.4              & +3.1              & 82.7            & 58.5             & +14.4              \\
DeRC                     & 86.4         & \textbf{78.1}          & \textbf{+4.6}          & 88.1         & \textbf{72.9}           & \textbf{+6.6}          & 89.2         & \textbf{60.5}          & \textbf{+20.4}          \\
\bottomrule[1pt]
\end{tabular}
\caption{Experimental results on two sets: 1) the validation sets of MNLI, FEVER, QQP; 2) their corresponding OOD test sets. Values of \textbf{$\Delta$} denote the performance gaps between debiasing methods and RoBERTa-base on the OOD test sets.
}
\label{tab:roberta_results}
\end{table*}
Afterwards, we develop DeRC and DePoE based on RoBERTa-base, and re-conduct experiments using the same hyperparameters as BERT-base. Experimental results are reported in Table \ref{tab:roberta_results}. Overall, DeRC still achieves the best performance on all OOD test sets, which proves that DeRC is also applicable to other Transformer-based models. 





\end{document}